\documentclass{sig-alternate-05-2015}

\usepackage{amsmath}
\usepackage{multirow}
\usepackage{listings}
\usepackage{hyperref}
\usepackage{array}
\usepackage{graphicx}
\usepackage{color}
\usepackage[labelfont=bf,belowskip=2pt,aboveskip=0pt]{caption}
\newcolumntype{L}[1]{>{\raggedright\let\newline\\\arraybackslash\hspace{0pt}}m{#1}}
\newcolumntype{C}[1]{>{\centering\let\newline\\\arraybackslash\hspace{0pt}}m{#1}}
\newcolumntype{R}[1]{>{\raggedleft\let\newline\\\arraybackslash\hspace{0pt}}m{#1}}

\begin{document}

% Copyright
\setcopyright{acmcopyright}

% DOI
%\doi{10.475/123_4}

% ISBN
%\isbn{123-4567-24-567/08/06}

%Conference
%\conferenceinfo{PLDI '13}{June 16--19, 2013, Seattle, WA, USA}

%\acmPrice{\$15.00}

%
% --- Author Metadata here ---
%\conferenceinfo{WOODSTOCK}{'97 El Paso, Texas USA}
%\CopyrightYear{2007} % Allows default copyright year (20XX) to be over-ridden - IF NEED BE.
%\crdata{0-12345-67-8/90/01}  % Allows default copyright data (0-89791-88-6/97/05) to be over-ridden - IF NEED BE.
% --- End of Author Metadata ---

%\title{Enhancing Yelp Reviews through Data Mining}
%\title{Is a Picture Worth Ten Thousand Words in a Review Dataset?}
\title{Is a Picture Worth Ten Thousand Words\\ in a Review Dataset?}
\numberofauthors{1} 

\author{	
\alignauthor
Roberto Camacho, Laura M. Rodriguez, Rebecca Urbina, and M. Shahriar Hossain\\
       	\affaddr{Dept. of Computer Science, University of Texas at El Paso, El Paso, TX 79968}\\
       \email{\normalsize{rcamachobarranco@utep.edu, \{lmrodriguez3, rurbina5\}@miners.utep.edu, mhossain@utep.edu}
       }
       %\hspace{3 mm}	
%% 1st. author
%\alignauthor
%Roberto Camacho\\
%       	\affaddr{University of Texas at El Paso}\\
%       	\affaddr{500 W. University Avenue}\\
%       	\affaddr{El Paso, TX 79968-0518}\\
%       \email{rcamachobarranco@utep.edu}\hspace{3 mm}
%% 2nd. author
%\alignauthor
%\alignauthor
% M. Shahriar Hossain\\
%       	\affaddr{University of Texas at El Paso}\\
%       	\affaddr{500 W. University Avenue}\\
%       	\affaddr{El Paso, TX 79968-0518}\\
%       \email{mhossain@utep.edu}
%\and
%% 3rd. author
%\alignauthor
%Laura M. Rodriguez\\
%       	\affaddr{University of Texas at El Paso}\\
%       	\affaddr{500 W. University Avenue}\\
%       	\affaddr{El Paso, TX 79968-0518}\\
%        \email{lmrodriguez3@miners.utep.edu}
%% 4th. author
%\alignauthor
%\alignauthor
%Rebecca Urbina\\
%       	\affaddr{University of Texas at El Paso}\\
%       	\affaddr{500 W. University Avenue}\\
%       	\affaddr{El Paso, TX 79968-0518}\\
%       \email{rurbina5@miners.utep.edu}
}

\maketitle

%%%%%%%%%%%%%%%%%%%%%%%%%%%%%%%%%%%%%%%%%%%%%%%%%%%%%%%%%%%%%%%%%%%%%%%%%%%%%%%%%%%%%%%%%%%%%%%%%%%%%%%%%%%
% Abstract: State the problem, then talk about the solution approach, then talk about the results
\begin{abstract}	
While textual reviews have become prominent in many re\-commendation-based systems, automated frameworks to provide relevant visual cues against text reviews where pictures are not available is a new form of task confronted by data mining and machine learning researchers. Suggestions of pictures that are relevant to the content of a review could significantly benefit the users by increasing the effectiveness of a review. We propose a deep learning-based framework to automatically: (1) tag the images available in a review dataset, (2) generate a caption for each image that does not have one, and (3) enhance each review by recommending relevant images that might not be uploaded by the corresponding reviewer. We evaluate the proposed framework using the Yelp Challenge Dataset. While a subset of the images in this particular dataset are correctly captioned, the majority of the pictures do not have any associated text. Moreover, there is no mapping between reviews and images. Each image has a corresponding business-tag where the picture was taken, though. The overall data setting and unavailability of crucial pieces required for a mapping make the problem of recommending images for reviews a major challenge. Qualitative and quantitative evaluations indicate that our proposed framework provides high quality enhancements through automatic captioning, tagging, and recommendation for mapping reviews and images. 

\end{abstract}


 
 %
%  Use this command to print the description
%


%%%%%%%%%%%%%%%%%%%%%%%%%%%%%%%%%%%%%%%%%%%%%%%%%%%%%%%%%%%%%%%%%%%%%%%%%%%%%%%%%%%%%%%%%%%%%%%%%%%%%%%%%%%%%%%%%%%%%%%%
% Introduction: Talk about the problem. Why this problem is hard. How you solve the problem. What are your contributions (itemize)?
%%%%%%%%%%%%%%%%%%%%%%%%%%%%%%%%%%%%%%%%%%%%%%%%%%%%%%%%%%%%%%%%%%%%%%%%%%%%%%%%%%%%%%%%%%%%%%%%%%%%%%%%%%%%%%%%%%%%%%%%

\section{Introduction} \label{sec:intro}

The usefulness of a review-based website (e.g., Yelp) largely depends on the quality of the materials produced by the reviewers. The heterogeneous nature of these materials provides a tremendous possibility to enhance user experience. For example, text reviews and images shared by many reviewers can be used to create snippets for users to quickly obtain a \textit{feeling} about the business. Karvonen, et al. \cite{karvonen2009widsets} show that visually prominent UI elements, such as images, play an important role in review-based decision making. However, production of mixtures of text and images is a difficult task for a review dataset due to the use of colloquial language, incorrect captioning of images, and insufficient labels for each of the images. Moreover, the images are captured %taken
by cameras of unknown configuration in uncontrolled environments, thus making extraction of image-features and mapping the features with textual units very challenging for any kind of enhancement. 
We propose a framework composed of a palette of deep learning and data mining techniques to recommend images for each review, even if a review was not originally submitted with a picture. In doing so, we predict tags for each image, generate captioning phrases for these images, and finally map reviews with the most relevant images. 
\begin{figure}[!b]
	\centering
	\includegraphics[width=1.0\columnwidth]{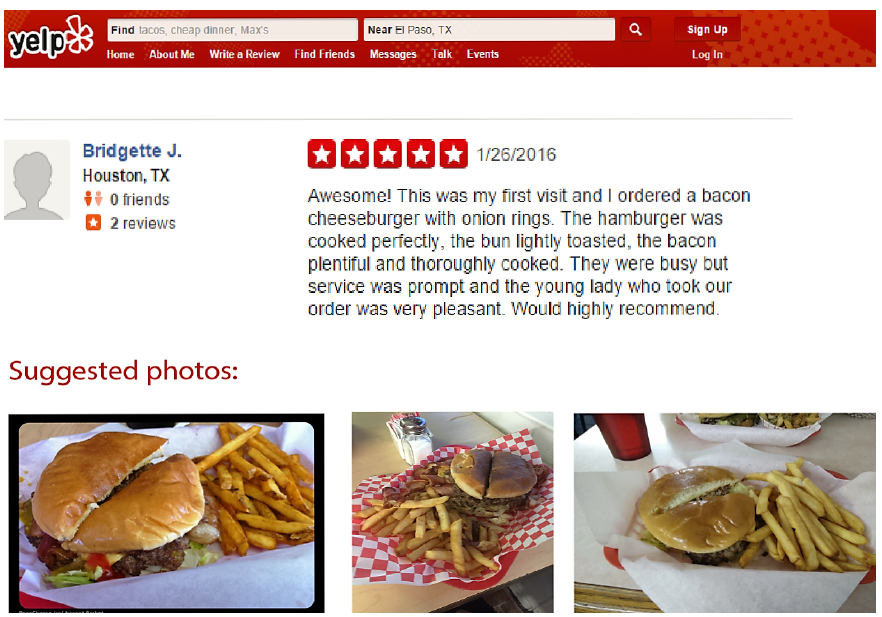}
	\caption{Suggested photos for a review describing a hamburger meal. \label{img:SuggestedPhotos}}
\end{figure} 
%The Yelp Dataset challenge invites participants to explore datasets made available by Yelp in order to explore innovative ways to use this data, in hopes of finding insight and correlations within the data that can help the business model grow. 
%Review-based websites, such as Yelp, Foursquare and TripAdvisor, usually depend on the quality of the material produced by its user base to provide a good experience for their end users. Thus, these websites usually try to identify the most helpful and reliable reviews using automated recommender systems. Karvonen, et al. show that visually prominent UI elements, such as images, play an important role in decision making for this type of websites~\cite{karvonen2009widsets}. Thus, reviews that are complemented with images will probably be more helpful for the user, and will have a higher influence on the user's choice. However, because reviews are user-generated, many of them do not include images that can depict the content of the review. The motivation of our project is to enrich the user experience and decision-making by developing a framework that automatically enhances Yelp reviews by suggesting images that are closely related to the latent topic of the review.

We leverage the data published by Yelp for the Yelp Dataset Challenge \cite{yelp_challenge}. The scope of this paper is limited to the review and photo datasets pertaining only to restaurants. The dataset contains 25,071 restaurants, 1,363,242 reviews of these restaurants, and 98,786 images. Many of the images do not have labels or captions, which are the primary links to connect reviews with images because the dataset does not provide an existing mapping between them. Our framework consists of three main components: 1) an image classifier used to predict the label of each image, 2) a captioning algorithm that generates possible captions for images that were not captioned by the reviewer, and 3) a mechanism to map a review to a number of most relevant images. As an outcome of the proposed framework, we will be able to recommend images for each review as shown in Figure~\ref{img:SuggestedPhotos}. %Figure~\ref{img:flow} presents the complete workflow of our framework.

%The main problem with this dataset is that many images are lacking labels and/or captions, which are the only way to link reviews to images. Thus, the framework developed consists of three main components: 1) an image classifier used to predict the label of each new image, 2) a captioning algorithm that generates possible captions for images that where not captioned by the user, and 3) an algorithm that allows us to match the most relevant words in a review to the most relevant words in a caption. Figure~\ref{img:flow} presents the complete workflow of our framework.

%\begin{figure*}[ht]
% \centering
%     \includegraphics[width=\textwidth]{images/flow}
% \caption{Framework Workflow\label{img:flow}}
%\end{figure*}

%The contributions of this paper are as follows.
%\begin{enumerate}
%	%\item We developed a deep-learning-based framework that suggests images that are relevant to the content of a review.
%	\item We describe a systematic approach to generate textual features for images.
%	\item We built and evaluated an image classification model using a CNN trained with the Yelp photos dataset.
%	\item We implemented and evaluated an image captioning model using an LSTM network.
%	\item Using the Yelp reviews dataset, we trained an LDA model and obtained the topic distribution for each review and image caption.
%\end{enumerate}

The paper contributes to the literature by describing a systematic approach to image recommendation for textual reviews, with minimal information available to create a mapping between both types. Section \ref{sec:problem_description} outlines the problem and Section \ref{sec:methodology} describes the overall framework. Section \ref{sec:evaluation} lists the evaluation techniques used. Section \ref{sec:experiments} provides descriptions of the experiments we performed. The related literature is described in Section \ref{sec:related_work} and we conclude the paper in Section \ref{sec:conclusions}.
 We deployed a Django-based website\footnote{Available at: \url{https://auto-captioning.herokuapp.com}} to visualize the outcomes.

% ACTIVE AGAIN

%%%%%%%%%%%%%%%%%%%%%%%%%%%%%%%%%%%%%%%%%%%%%%%%%%%%%%%%%%%%%%%%%%%%%%%%%%%%%%%%%%%%%%%%%%%%%%%%%%%%%%%%%%%%%%%%%%%%%%%%
% Problem description:This is where you provide the formal description of the problem.
\section{Problem Description}\label{sec:problem_description}
Let $D = \{(i_1,l_1,c_1,b_1),(i_2,l_2,c_2,b_2),...,(i_N,l_N,c_N,b_N)\}$ be an image dataset containing $N$ images ($i$), along with their corresponding labels ($l$), captions ($c$), and business id ($b$). A label can be a value from the following set of categories: \{\texttt{food}, \texttt{inside}, \texttt{outside}, \texttt{drink}, \texttt{menu}, \texttt{none}\}. A caption is expected to be a sentence but can be empty as well. Each image has exactly one business id. 

Let $R = \{r_1, r_2,...,r_{|R|}\}$ be a text dataset which contains all the reviews of a specific restaurant $b$. We seek for a mapping $M$, such that given a review $r \in R$ we can select top $k$ images $(i, l, c, b)\in D$ such that the image $i$ is closely related to review $r$. To establish such a mapping, we rely on relationships between a caption $c$ of an image and review $r$. 
Thus, a major task in the proposed framework is to generate a caption $c$ for image $i$, in case one does not exist, which can then be used to relate $i$ to review $r$. In turn, generating a caption $c$ is performed using the probabilities of $i$ belonging to the different possible categories of $l$. 
In summary, we have three subtasks: (1) generate a label for images with \texttt{none} label, (2) generate caption-words based on the labels and the subset of images that has captions, and (3) for each review, find a set of relevant images by topically comparing the review and the image captions.

%From the Yelp dataset, we know that all values of $l$ belong to the set $L=\{food, drink, inside, outside, menu, none\}$ and that there exist tuples within $D$ with empty captions, i.e. $c=``"$. When an image does not contain a caption, it is not possible to map this image to a review. Thus, we need a way to generate a caption based on an image. Using a supervised image classification algorithm, we can find a function that models the relation between all images and their labels. This function can be used to predict the most probable label for an image $i$ labeled as ``none'', or labeled incorrectly. Furthermore, we can use this function along with the image information to generate a new caption $c$. Once a caption is available, it is possible to verify if this caption is highly related to review $r$.

%In summary, we have three different problems: (1) find images that are highly related to reviews by comparing the image captions with the reviews; (2) generate a caption based on an image and its label distribution when a caption is not available; and (3) generate a label based on an image when the label is ``none''.

%%%%%%%%%%%%%%%%%%%%%%%%%%%%%%%%%%%%%%%%%%%%%%%%%%%%%%%%%%%%%%%%%%%%%%%%%%%%%%%%%%%%%%%%%%%%%%%%%%%%%%%%%%%%%%%%%%%%%
% Methodology:Describe the entire solution system. Will contains many subsections.
\section{Methodology}\label{sec:methodology}
Our framework solves the problem of recommending images for each review in three major steps. First, we use an image classifier to predict a label for images categorized as \texttt{none}. Second, we use a captioning algorithm, with the image features (obtained in the first step) and existing captions as inputs. This generates caption-words for images that do not have captions. Finally, we apply topic modeling on the reviews and captions separately to be able to create a probabilistic mapping. The following subsections describe these steps.
%Figure~\ref{img:flow} presents the complete workflow of our framework.
%The main objective of the proposed framework is to automatically enhance Yelp reviews by suggesting images that are closely related to relevant terms mentioned in the reviews. The proposed framework consists of three main components. First, we use an image classifier to predict the label $l\in\{food, drink, inside, outside, menu\}$ for images labeled as ``none''. Second, we use a captioning algorithm, with the image features and actual captions as inputs, which generates descriptions for images that were not captioned by the user. Finally, we apply a topic modeling algorithm on the review and caption datasets, which include the generated captions in case they were not provided. We suggest relevant images for a particular review by matching captions and reviews based on the top words from the most relevant topic of the enhanced review. Figure~\ref{img:flow} presents the complete workflow of our framework.

%%%%%%%%%%%%%%%%%%%%%%%%%%%%%%%%%%%%%%%%%%%%%%%%%%%%%%%%%%%%%%%%%%%%%%%%%%%%%%%%%%%%%%%%%%%%%%%%%%%%%%%%%%%%%%%%%%%%%%%%
\subsection{Image classification} \label{sec:class}
The first step in our framework is to categorize each of the images labeled as \texttt{none} to one of the following categories: \texttt{food}, \texttt{inside}, \texttt{outside}, \texttt{drink} or \texttt{menu}. Our preliminary data analysis reveals that more than 25\% of the restaurant images are labeled as \texttt{none}. We use a Convolutional Neural Network (CNN) image classifier to obtain class probabilities for all images in the test set (labeled as \texttt{none}). 

Convolutional Neural Network algorithms require that all of the images have the same dimension and are shaped as a square. We resized the images so that the smallest dimension of the image is 64 or 224 pixels, and then cropped the image in the other dimension to obtain a 64-by-64-pixel or 224-by-224-pixel image. We tested and implemented CNN models using two Python 2.7 libraries based on the Theano deep-learning library~\cite{Bastien-Theano-2012}: Keras~\cite{chollet2015}, and Lasagne~\cite{sander_dieleman_2015_27878}. Keras and Lasagne provide high-level functions for deep learning algorithms, including convolution, pooling and fully-connected layers, as well as backpropagation and optimization routines, whereas Theano provides the back-end of the computation and includes GPU support.
%\robc{Theano is also a library :)}

We used a number of different CNN models to evaluate their accuracy. One of these models was based on the CIFAR10 data~\cite{chollet2015} while the others were designed to work with the ImageNet data~\cite{deng2009imagenet}: VGG-16~\cite{simonyan2014very}, VGG-19~\cite{simonyan2014very} and GoogleNet~\cite{szegedy2015going}. The CIFAR10 model is relatively simple, with only 11 layers. The VGG-16 model adds four convolutional layers and one fully-connected layer, which significantly increases the complexity of the model. The VGG-19 and GoogleNet models add even a larger number of layers, consisting of 19 and 22, respectively. We also used MATLAB\textsuperscript{\textregistered}'s Bag-of-Features with SVM classification algorithm as a baseline. We used six-fold cross validation for evaluation of all these approaches.
\subsection{Image captioning} \label{sec:captioning}
\begin{figure}[!b]
	\centering
	\includegraphics[width=0.5\textwidth]{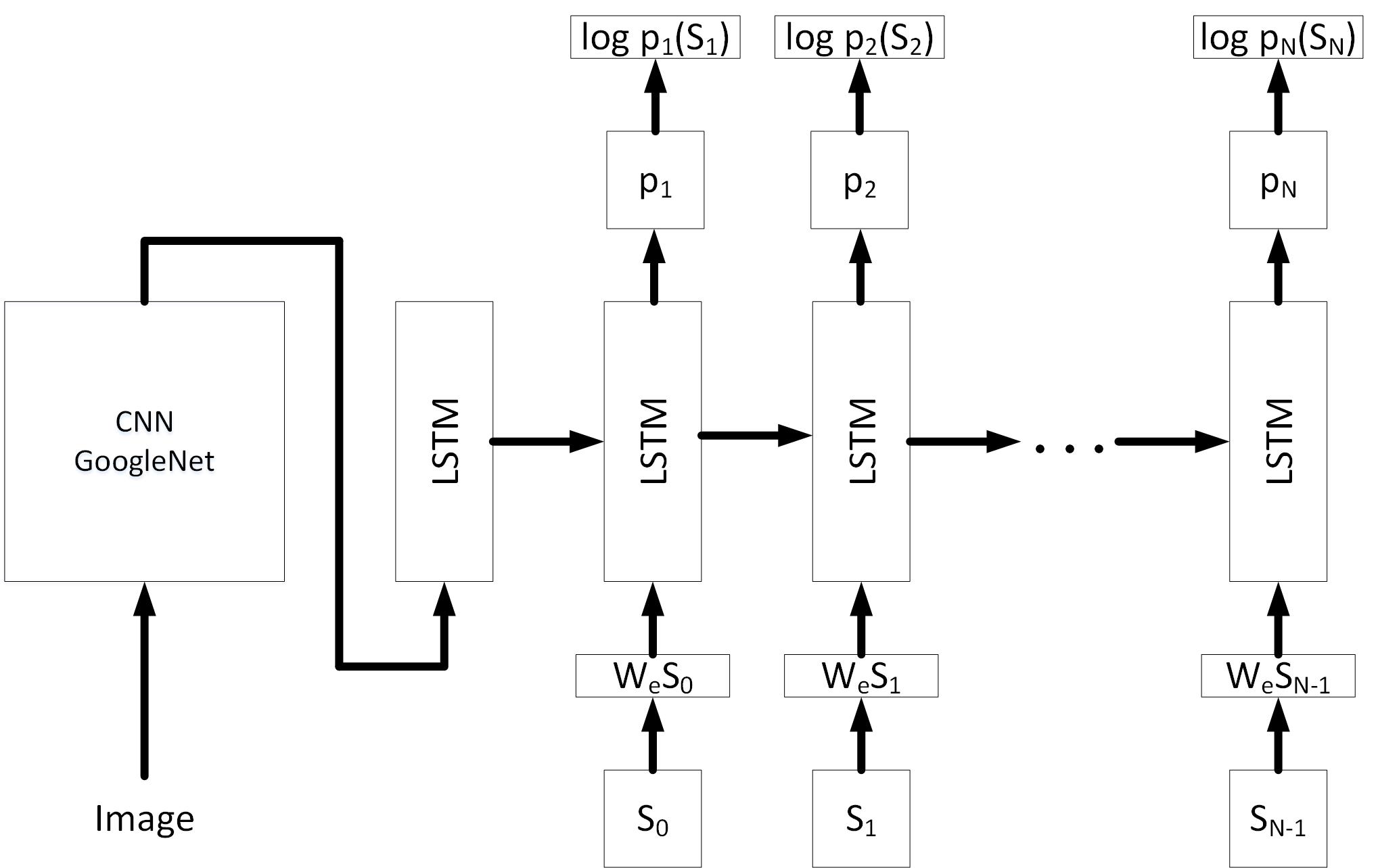}
	\caption{Structure of the Neural Image Caption (NIC) generator algorithm.\label{img:nic}}
\end{figure}

\begin{figure*}[!t]
	\centering
	\includegraphics[width=1.0\textwidth]{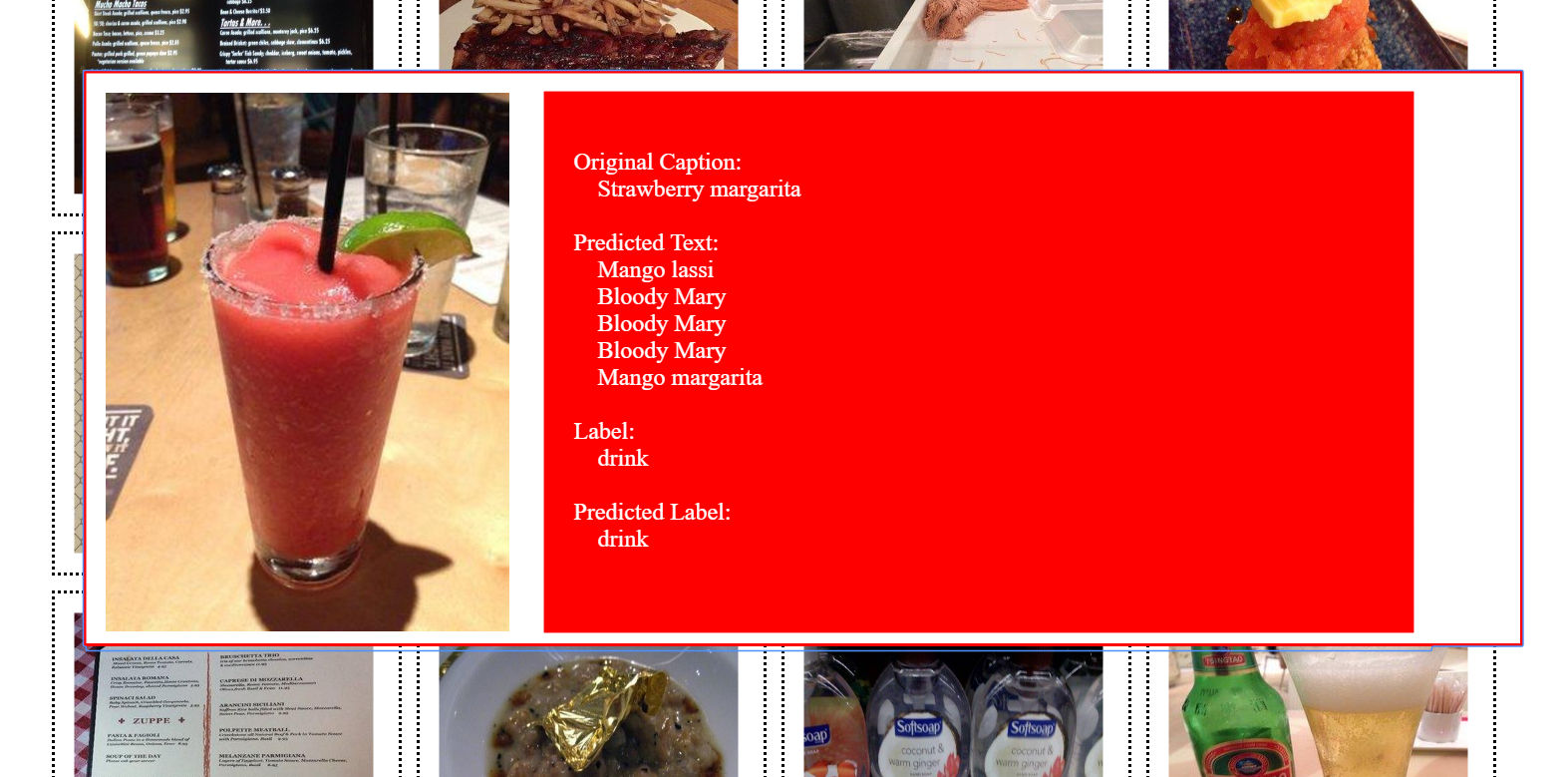}
	\caption{A screenshot of our website showing predicted caption-words for an image of a beverage. \label{img:LabelCapSuggestions}}
\end{figure*}
We leverage a Lasagne-based implementation of the Neural Image Caption (NIC) generator \cite{vinyals2015show} to predict captions for images with no caption. The NIC generator uses a special form of a recurrent neural network (RNN) called Long-Short Term Memory (LSTM) network to sequentially create a fixed-dimensional vector, required due to the variable length of the input and output sentences. LSTM nets are a special type of RNN capable of learning long-term dependencies. LSTM nets apply weighted layered gates between the input, output and previous hidden states. By assigning different magnitudes to every gate (between 0 and 1), the information flow is modified so that the previous information is useful for the model, and if not, the model \textit{forgets} the information. LSTM nets are able to train the gates automatically through backpropagation, obtaining a more robust model with higher accuracy~\cite{graves2005framewise}.
%In this work, we use a Lasagne-based implementation of the algorithm presented by Vinyals, et al. in~\cite{vinyals2015show} called Neural Image Caption (NIC) generator. The NIC generator uses a special form of a recurrent neural network (RNN) called Long-Short Term Memory (LSTM) network to sequentially create a fixed-dimensional vector, required due to the variable length of the input and output sentences~\cite{vinyals2015show}. LSTM networks (nets) are a special type of RNN capable of learning long-term dependencies. LSTM nets apply weighted layered gates between the input, output and previous hidden states. By assigning different magnitudes to every gate (between 0 and 1), the information flow is modified so that the previous information is actually useful for the model, and if not, it ``forgets'' it. LSTM nets are able to train these gates automatically through backpropagation, obtaining a more robust model with higher accuracy~\cite{graves2005framewise}.
 
The LSTM net uses information about an image as input. We obtain a lower-dimensional representation for each image using a Convolutional Neural Network (CNN). Out of the several CNN models described in Section~\ref{sec:class}, we chose the GoogLeNet model trained on 224-by-224 images to feed the image features to the LSTM net because of GoogLeNet's flexible compatibility with LSTM. The captioning results presented in this paper are resultant from image features generated by GoogLeNet. The complete model is outlined in Figure~\ref{img:nic}. In the figure, $S_i$ represents a word of a sentence, $W_e$ represents the trained parameters, $p_i$ is a probability distribution over all the words in the vocabulary and the $\log p_i(s_i)$ is the log-likelihood of the correct word at each step.

Captions are generated by maximizing the log-likelihood of the probability of obtaining the correct caption given an image. The LSTM net is trained sequentially and takes into account the image as well as all of the previously seen words to infer the next word of the output sentence. At each position of the output sentence only the word with the highest probability is selected, which has the disadvantage of not providing the globally optimal solution. The detailed sampling method required for this inference model is described in~\cite{vinyals2015show}.

While prediction of caption-words by learning relationships using existing captions seems a logical direction, we did not target the problem of correcting captions in case they are not appropriate. Based on our study, not reported in this paper, many of the captions do not describe the relevant image well. For example, a caption ``Hurrah it's my birthday'' for a picture of a pasta dish only increases the noise-level rather than providing informative features. A possible solution is crowdsourcing a subset of data to obtain an accurate training set. However, this aspect is out of the scope of the current paper.

Figure \ref{img:LabelCapSuggestions} shows a screenshot of our website for a picture of a glass of margarita. The suggested caption-words include margarita as well as another cocktail, Bloody Mary. This sample indicates that our proposed system was able to closely predict content of the image and map them with textual snippets. %\robc{Moved this here, since it is talking about the captioning part}

\subsection{Topic modeling and review enhancement} \label{sec:topic}
We leverage Latent Dirichlet Allocation (LDA) \cite{Blei:2003:LDA:944919.944937} to model the topics of the reviews. For each review, we select the best topic and select the top $t$ representative terms of that topic, regardless if they appear on the review or not. For each review, we recommend the top $\phi$ images based on the presence of the $t$ representative terms in the review and in the captions of the images available for the business for which the review was written. An image is ranked higher for a particular review if a representative term is present both in the image caption and in the review, compared to an image which contains the representative term only in its caption. 
%\robc{Removed the following: but the review does not contain the term}
% but the review does not contain the term. 

We start by selecting images using representative terms that are present in both the review and the image caption. If $\phi$ images cannot be found, we select images for which captions contain representative terms but the review does not. This process ensures that the image selection is not solely driven by overlaps between a review and a caption, rather reviews and image captions without any overlap may become candidates for potential mapping due to the use of topical terms during the ranking. Figure~\ref{img:SuggestedPhotos} shows a sample of recommended images for a text review written for a burger.

\section{Evaluation}\label{sec:evaluation}
We use different metrics to evaluate the quality of the results for each of the main components of the framework: image classification, image captioning, and topic modeling. For image classification, we use the top-1 accuracy, which is the percentage of test images that were classified correctly, as defined by Equation~\ref{eq:accuracy}. We only use the top-1 accuracy because of the small number of possible labels. 
\begin{equation} \label{eq:accuracy}
accuracy = \frac{images\ labeled\ correctly}{total\ images}*100\% 
\end{equation}

The evaluation of the quality of the image captioning results was performed using a combination of two different metrics: Bilingual Evaluation Understudy (BLEU) and a confidence score. The BLEU method, proposed by Panineni et al.~\cite{papineni2002bleu}, computes the geometric mean of n-gram precisions. Since the training set consists mostly of short sentences (captions), we removed the brevity penalty typically used, which prevents that very short sentences have very high scores with just a few words match. We obtained a range of BLEU-1 to BLEU-4 scores using the \texttt{corpus\_bleu} function of NLTK, a Python library focused on Natural Language Processing. The BLEU metric has some shortcomings, particularly because it does not take into account the probability with which a caption is generated. 

Because of this limitation, we designed a metric that measures a confidence score for each generated caption. When a caption is being generated, we take the top $k$ candidate words at each position in a sentence. For each position, we use Equation~\ref{eq:nonuniformity} to measure the non-uniformity of probabilities for the $k$ candidates: 
\begin{equation} \label{eq:nonuniformity}
\nu(X) = \frac{||U(\frac{1}{k})-X||_1}{2-2/k}
\end{equation}
where $X$ is the normalized probability distribution of the top $k$ candidates where $k>1$, and $U$ is a uniform probability distribution over size $k$. Ideally, we would like a very high non-uniformity value ($\nu(X)=1$) which means that we are very confident that the top word should be next. A uniform distribution ($\nu(X) = 0$) would indicate a random selection of top words. We compute the confidence for a generated caption with the following equation:
\begin{equation}
Confidence(W) = \frac{\left[\sum^m_{i=1}(e^{\nu(v(w_i))*p(w_i)})\right]-m}{[m*e^1]-m}
\end{equation}
where $W$ is an array of $m$ words, $W=\{w_1, w_2,...,w_m\}$ representing the generated sentence, $v(w)$ is the normalized truncated probability distribution of top $k$ words for each position in the sentence in which $w$ has the highest probability, and $p(w)$ is the probability of the word $w$ from the original distribution. This metric has a range of $[0.0, 1.0]$, where higher confidence is better.

The evaluation of the LDA topics was performed using perplexity, which measures the model fit of an unseen set of documents, where the value decreases as a function of the log-likelihood of the holdout documents. An LDA model with lower perplexity is better. We use the following bounded definition of perplexity, presented by Hoffman, et al.~\cite{hoffman2010}:
\begin{equation}\label{eq:perplexity}
\begin{split}
perplexity(n^{test},\lambda,\alpha)\leq exp\lbrace -(\sum_i{\mathbb{E}_q[\log p(n_i^{test}, \theta_i,z_i|\alpha,\beta)]} \\
-\mathbb{E}_q[\log q(\theta_i,z_i)])(\sum_{i,w}{n^{test}_{iw}})\rbrace
\end{split}
\end{equation}
Where $n^{test}$ is the total word count of the holdout documents, $\lambda$ is the posterior parametrization over $\beta$, $\alpha$ is the Dirichlet parameter, $n^{test}_i$ is the word count of holdout document $i$, $\theta_i$ is a vector of topic weights for document $i$, $z_i$ is a vector of per-word topic assignments for the words in document $i$, $\beta$ is a dictionary of topics, $q$ is a variational distribution which is indexed by a set of free parameters, and $n_{iw}^{test}$ is the number of times the word $w$ appears in the holdout document $i$.

%%%%%%%%%%%%%%%%%%%%%%%%%%%%%%%%%%%%%%%%%%%%%%%%%%%%%%%%%%%%%%%%%%%%%%%%%%%%%%%%%%%%%%%%%%%%%%%%%%%%%%%%%%%%%%%%%%%%%%%%
% Experiments:Itemize a number of questions you seek to answer. For every question, write a subsection with a relevant experiment.
\section{Experiments}\label{sec:experiments}
In this section, we seek to answer the following questions about the proposed framework.
\begin{enumerate}
  \setlength{\itemsep}{1pt}
  \setlength{\parskip}{0pt}
  \setlength{\parsep}{0pt}	
	\item How do different CNN architectures compare in terms of the accuracy of image classification? How do these results compare with a simpler method, e.g. SVM using bag-of-features? (Section~\ref{sec:results_classification})
	\item How does changing model hyperparameters for the NIC generator affect the results in terms of confidence and BLEU-4 score? What is the relationship between the confidence score and the BLEU score for the generated captions? (Section~\ref{sec:results_captioning})
	\item How well does our framework generate captions for the images of the Yelp dataset? (Section~\ref{sec:results_users})
	\item How does varying different hyperparameters affect the perplexity of the resulting LDA model? (Section~\ref{sec:results_topics})
	\item Can we relate reviews with images based on the top words obtained through topic modeling? (Section~\ref{sec:results_reviews})
\end{enumerate}
%In Section~\ref{sec:platform} we describe the experimental platform used for our experiments. 
We focused on the 25,071 (out of 77,445) restaurants detected in the Yelp dataset. The restaurant image dataset contains 98,786 images with an average of 4 images per restaurant; 62\% of the images do not contain any caption and 25\% are labeled as \texttt{none}. The total number of reviews for restaurants is 1,363,242.
%%%%%%%%%%%%%%%%%%%%%%%%%%%%%%%%%%%%%%%%%%%%%%%%%%%%%%%%%%%%%%%%%%%%%%%%%%%%%%%%%%%%%%%%%%%%%%%%%%%%%%%%%%%%%%%%%%%%%%%%
%\subsection{Experimental platform} \label{sec:platform}

Deep Neural Networks, such as CNNs and LSTM nets, require large memory and computing-power. We used two different devices for our experiments: an ASUS K501UX laptop, and the Griffin computing cluster, at the University of Texas at El Paso. The ASUS laptop has a 2.5 GHz Intel Core i7 6500U processor, 8GB memory, and NVIDIA GTX950M 914 MHz 2GB video card. Due to the low memory capacity of the NVIDIA GTX950M, the laptop was limited to training images of 64 by 64 pixels. Table~\ref{tbl:griffin} provides the configuration of the Griffin computing cluster.

\begin{table}[t!]
\centering
\caption{Characteristics of the Griffin cluster.}
\label{tbl:griffin}
\begin{tabular}{|l|l|}
\hline
\multicolumn{1}{|c|}{\textbf{Component}} & \multicolumn{1}{c|}{\textbf{Technology}} \\ \hline
Nodes                                                                          & (7x) Compute Nodes                                                             \\ \hline
                                                                               & (2x) AMD Opteron 6220                                                         \\ \cline{2-2} 
                                                                               & 8 cores/processor                                                              \\ \cline{2-2} 
\multirow{-3}{*}{Processor}                                                    & 1 thread/core                                                                  \\ \hline
Memory (RAM)                                                                   & 4GB/core                                                                       \\ \hline
Hard Drive                                                                     & 1TB SATA 7.2K RPM  16MB cache                                                  \\ \hline
                                                                               & (2x) NVIDIA Tesla M2090, 1.3GHz                                                \\ \cline{2-2} 
                                                                               & 512 CUDA cores                                                                 \\ \cline{2-2} 
\multirow{-3}{*}{Video Card}                                                   & 6GB GDDR5                                                                      \\ \hline
\end{tabular}
\end{table}

%%%%%%%%%%%%%%%%%%%%%%%%%%%%%%%%%%%%%%%%%%%%%%%%%%%%%%%%%%%%%%%%%%%%%%%%%%%%%%%%%%%%%%%%%%%%%%%%%%%%%%%%%%%%%%%%%%%%%%%%
\subsection{Image classification} \label{sec:results_classification}
We tested different CNN architectures to compare the accuracy obtained with each of them for the image classification problem. The architectures include: a simple CIFAR10 model (11-layer deep), VGG-16~\cite{simonyan2014very} (16-layer deep), VGG-19~\cite{simonyan2014very} (19-layer deep) and GoogleNet~\cite{szegedy2015going} (22-layer deep). We also used the MATLAB\textsuperscript{{\textregistered}} Bag-of-Features-based SVM classifier as a baseline, to verify the gain in accuracy obtained by using a neural network versus a simpler method. 

All of the models were used to classify the images into five different classes. The weight initialization for all the CNN models was a uniform random distribution between $-0.5$ and $0.5$. Table~\ref{tbl:cnn_results} presents the configurations used and the accuracy obtained for each model. The results indicate that using neural networks improves significantly the accuracy of the image classification models. The models using 64-by-64-pixel images were trained from scratch using Keras, while the models using 224-by-224-pixel images were trained with Lasagne after initializing the model weights to the ones obtained using the ImageNet dataset. This, along with the smaller learning rate and the higher number of epochs trained, might explain the higher accuracy of the models trained with 64-by-64-pixel images.

% \begin{table}[]
% \centering
% \caption{CNN Models Tested}
% \label{tbl:cnn_models}
% \begin{tabular}{|l|l|l|}
% \hline
% \rowcolor[HTML]{000000} 
% {\color[HTML]{FFFFFF} Model} & {\color[HTML]{FFFFFF} Image size} & {\color[HTML]{FFFFFF} Learning rate} \\ \hline
% Simple model                 & 64 x 64                           & 0.0001                               \\ \hline
% VGG-16                       & 64 x 64                           & 0.0001                               \\ \hline
% VGG-16                       & 224 x 224                         & 0.001                                \\ \hline
% VGG-19                       & 224 x 224                         & 0.001                                \\ \hline
% GoogleNet                    & 224 x 224                         & 0.001                                \\ \hline
% \end{tabular}
% \end{table}

\begin{table}[t!]
\centering
\caption{Results for image classification based on labels.}
\label{tbl:cnn_results}
\begin{tabular}{|l|r|r|r|r|}
\hline
\multicolumn{1}{|c|}{\textbf{Model}} & \multicolumn{1}{C{1.35cm}|}{\textbf{Image size}}& \multicolumn{1}{C{1.05cm}|}{\textbf{Learn. rate}} & \multicolumn{1}{c|}{\textbf{Epochs}} & \multicolumn{1}{C{1cm}|}{\textbf{Accy.}} \\ \hline
Simple model & 64 x 64 & 0.0001 & 160 & 94.12\% \\ \hline
VGG-16 & 64 x 64 & 0.0001 & 160 & 94.78\% \\ \hline
VGG-16 & 224 x 224 & 0.001 & 10 & 83.79\% \\ \hline
VGG-19 & 224 x 224 & 0.001 & 10 & 78.93\% \\ \hline
GoogleNet & 224 x 224 & 0.001 & 10 & 88.21\% \\ \hline
BoF SVM & 64 x 64 & N/A & N/A & 69.00\% \\ \hline
\end{tabular}
\end{table}

\begin{table}[t!]
	\centering
	\caption{Detailed results for image classification using images of 64 by 64 pixels.}
	\label{tbl:results_64}
	\begin{tabular}{|l|c|c|}
		\hline
		{} & {\textbf{CIFAR10 model}} & {\textbf{VGG-16 model}} \\ \hline
		\multicolumn{1}{|l|}{Epochs}                                                                 & \multicolumn{2}{c|}{160 epochs}                                                              \\ \hline
		\multicolumn{1}{|l|}{Image size}                                                             & \multicolumn{2}{c|}{64 x 64 pixels}                                                          \\ \hline
		\multicolumn{1}{|l|}{\begin{tabular}[c]{@{}l@{}}Average time\\ (sec per epoch)\end{tabular}} & \multicolumn{1}{c|}{188}                      & \multicolumn{1}{c|}{1,220}                   \\ \hline
		\multicolumn{1}{|l|}{Total time}                                                             & \multicolumn{1}{c|}{$\sim$8 hours}            & \multicolumn{1}{c|}{$\sim$54 hours}          \\ \hline
		\multicolumn{1}{|l|}{\begin{tabular}[c]{@{}l@{}}Best training \\ accuracy\end{tabular}}      & \multicolumn{1}{c|}{95.70\%}                  & \multicolumn{1}{c|}{99.78\%}                 \\ \hline
		\multicolumn{1}{|l|}{\begin{tabular}[c]{@{}l@{}}Best test \\ accuracy\end{tabular}}          & \multicolumn{1}{c|}{94.12\%}                  & \multicolumn{1}{c|}{94.78\%}                 \\ \hline
	\end{tabular}
\end{table}

\begin{figure}[b!]
	\centering
	\includegraphics[width=0.5\textwidth]{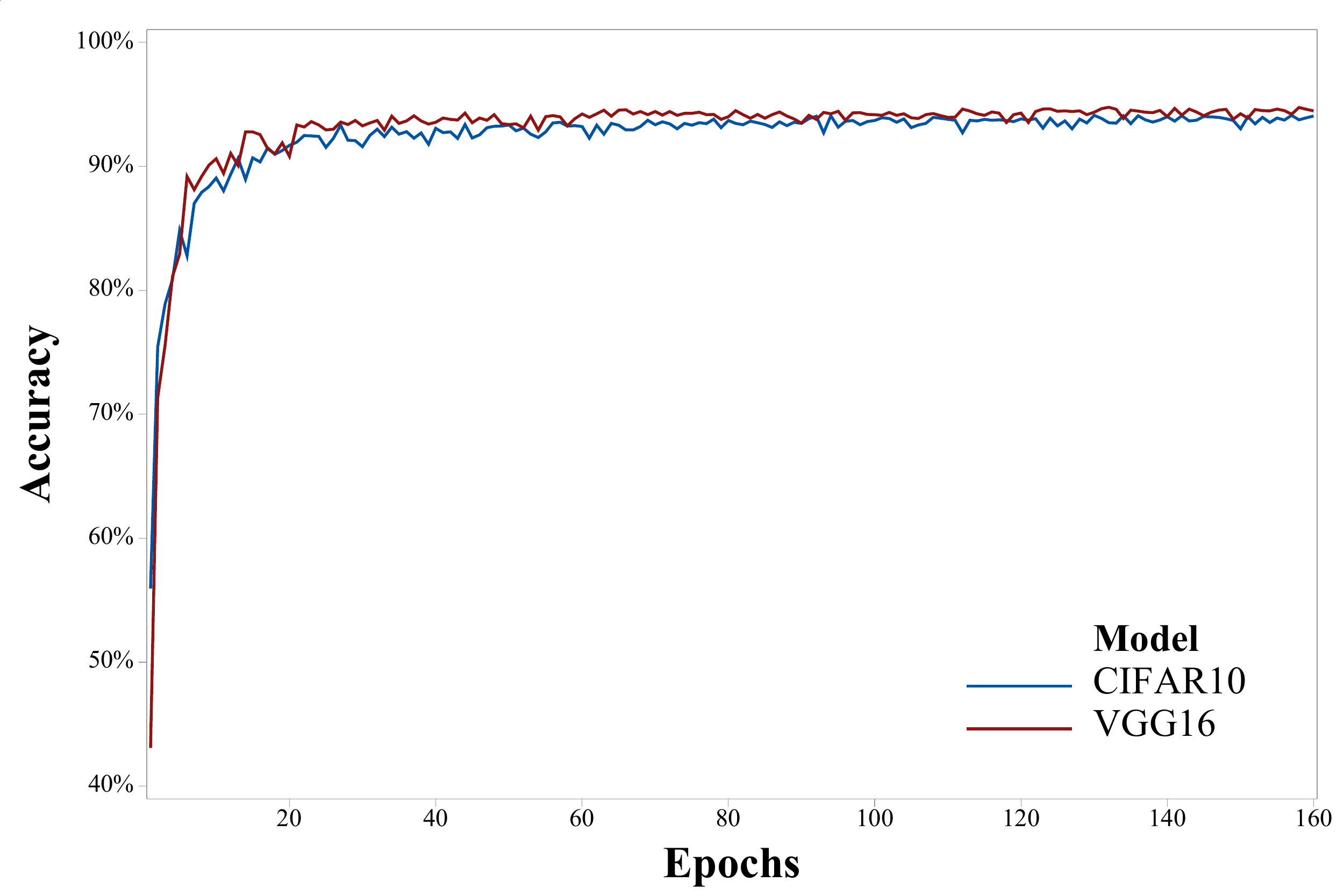}
	\caption{Accuracy per epoch on testing data.\label{accuracy}}
\end{figure}

Table~\ref{tbl:results_64} presents more details about the results for the models that use images of 64 by 64 pixels. Each of these models was trained for 160 epochs. The VGG-16 model obtained 94.78\% accuracy on the test data, while the CIFAR10 model obtained a 94.12\% accuracy. Our observation is that the difference in the accuracy for the training data is significantly larger, and that the error on this dataset for the VGG-16 model is very small (0.22\%). Thus, the VGG-16 model might be overfitting the training data.

The VGG-16 model took significantly longer to train compared to the simple model, with 54 hours against the 8 hours that the CIFAR10 model required. This is a result of the difference in the number of layers, and in particular the number of convolution operations required by each model. While the CIFAR10 model only has one convolutional layer, the VGG-16 has five. 

Figure~\ref{accuracy} shows how the accuracy for both of the models changes as the number of epochs increases. As can be seen, the CIFAR10 model outperforms the VGG-16 for the first five epochs. After the sixth iteration, the VGG-16 model appears to be consistently better than CIFAR10, albeit by a small margin. The graph also shows that the accuracy of each classifier seems to reach a steady state, but the accuracy still varies from epoch to epoch, which indicates that the training may be reduced to around 60 epochs.

\begin{figure}[b!]
	\centering
	\includegraphics[width=0.5\textwidth]{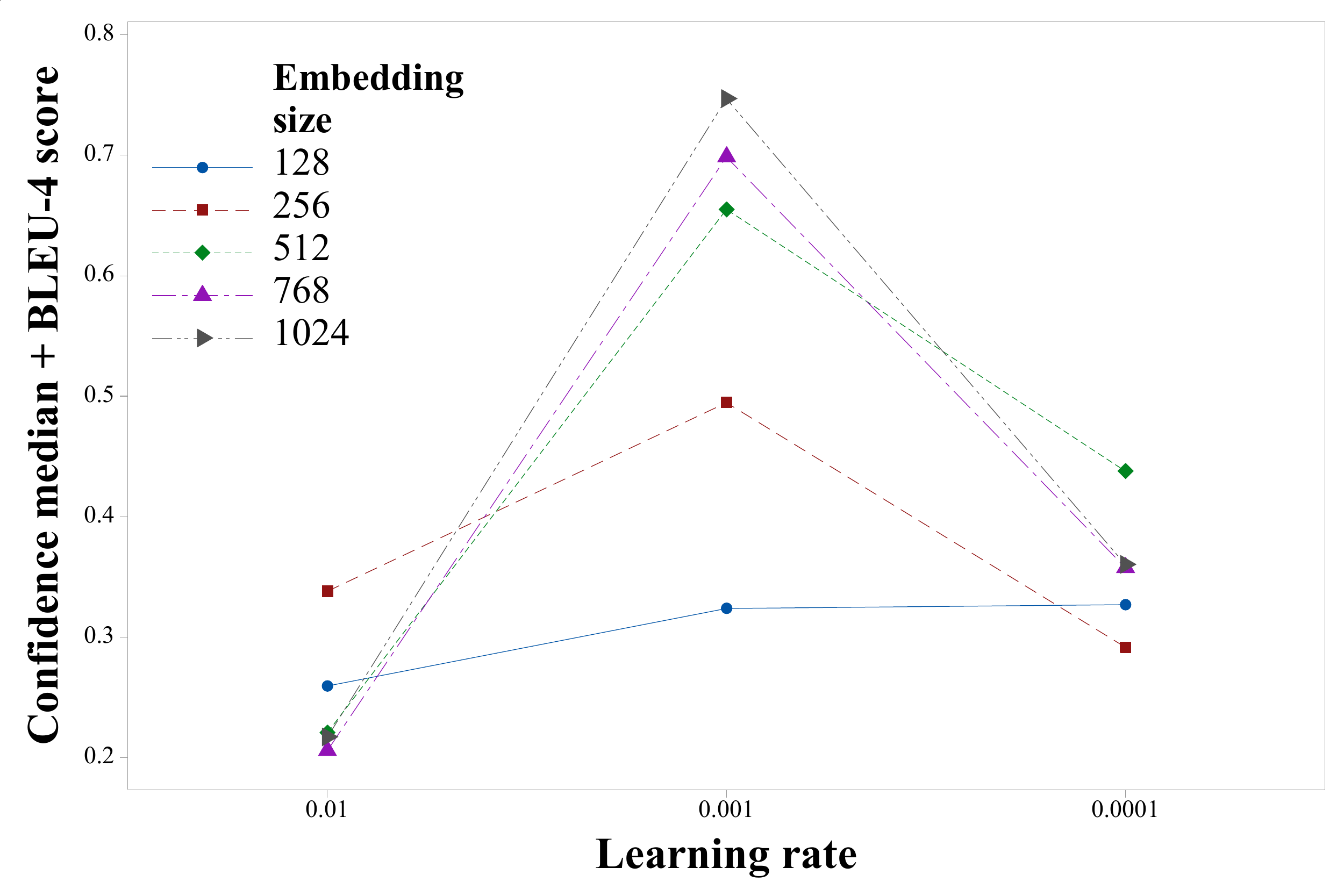}
	\caption{Effect of embedding size and learning rate settings on quality of NIC generator.\label{fig:lstm_exp1}} %TODO
\end{figure}
 
To further evaluate the CNN model used for the image captioning component of our framework, GoogLeNet, we performed a participant-based evaluation of the predicted labels of 240 randomly selected images originally labeled as \texttt{None}. All these images and the predicted labels were given to one participant for evaluation. The participant was asked to mark the predicted labels as \textit{correct} or \textit{incorrect} and comment on anything observed. Based on the participant's comments, there were 38 images that did not belong to any of the five classes because of lack of relevance of the image contents with the labels. These 38 images include phone numbers, face images, group photos, and other irrelevant images.
%\rob{because of lack of relevance of the image content with respect to the labels. These 38 images include phone numbers, face images, group photos, and other irrelevant images.} 
Out of the 202 remaining images, according to the participant, 149 images were labeled correctly (73.76\%). Our observation from this study was that the CNN model did not perform as good when attempting to distinguish between \texttt{inside} and \texttt{outside} labels, particularly in dark settings. We proceed with the results obtained from GoogLeNet because this model provides a decent level of accuracy on average and is compatible with the neural image caption generator.
 
%\rob{To further evaluate the CNN model used for the image captioning component of our framework, i.e. GoogLeNet, we performed a manual evaluation of the predicted labels of 240 random images originally labeled as \texttt{None}. Out of the 240 images, there were 38 images that did not belong to any of the five classes used by Yelp and were removed from the evaluation. Out of the 202 remaining images, 149 images were labeled correctly (73.76\%). The CNN model had the most errors when trying to distinguish between \texttt{inside} and \texttt{outside} labels, particularly for dark settings. Thus, we consider the performance of the model acceptable.}

%%%%%%%%%%%%%%%%%%%%%%%%%%%%%%%%%%%%%%%%%%%%%%%%%%%%%%%%%%%%%%%%%%%%%%%%%%%%%%%%%%%%%%%%%%%%%%%%%%%%%%%%%%%%%%%%%%%%%%%%
\subsection{Image captioning} \label{sec:results_captioning}
The NIC generator %model we used for captioning has two inputs: (1) the image features from the CNN model and (2) the image captions from the Yelp dataset. There are 
has several hyperparameters that can affect the quality of the resulting captions: maximum sequence length (\texttt{max\_seq\_len}), batch size (\texttt{batch\_sz}), embedding size (\texttt{embedding\_sz}), learning rate (\texttt{lr}), and number of iterations (\texttt{iters}). To select the hyperparameters that resulted in the best confidence and BLEU-4 score, we created an initial suite of experiments setting \texttt{max\_seq\_len} to 6, \texttt{batch\_sz} to 50 and \texttt{iters} to 20,000. Figure~\ref{fig:lstm_exp1} depicts the results of setting \texttt{embedding\_sz} to \{128, 256, 512, 768, 1024\} and \texttt{lr} to \{0.01, 0.001, 0.0001\}. The Y-axis represents the maximum value, across the iterations of one model, of adding the median of the confidence scores for all the generated sentences to the corpus-level BLEU-4 score. Thus, the hypothetical maximum value in this dimension is 2.0. In this case, it is clear that the optimal result is obtained when \texttt{embedding\_sz} is 1,024 and \texttt{lr} is 0.001.

\begin{figure}[t!]
	\centering
	\includegraphics[width=0.5\textwidth]{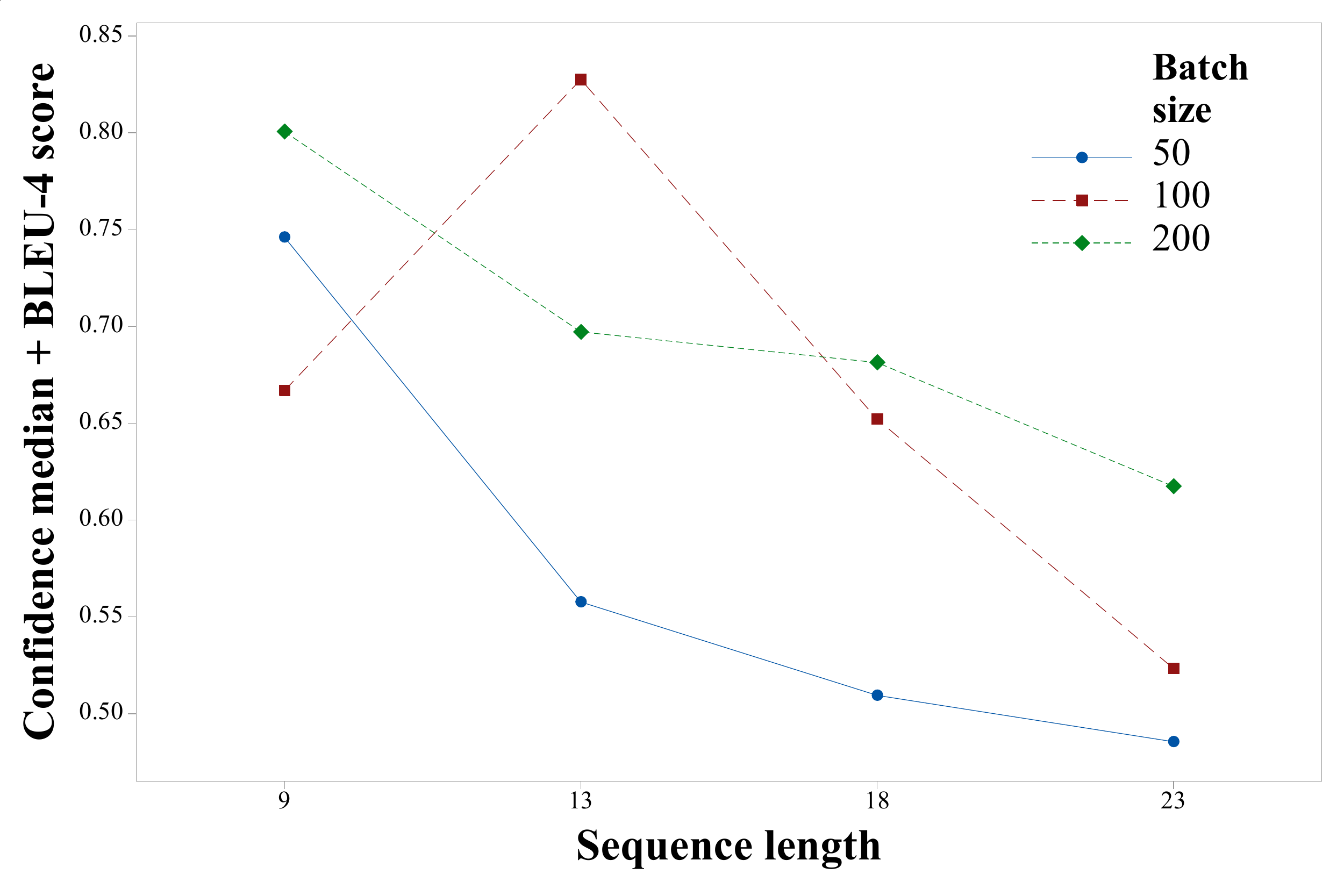}
	\caption{Effect of sequence length and batch size settings on quality of NIC generator.\label{fig:lstm_exp2}} %TODO
\end{figure}

\begin{figure}[b!]
	\centering
	\includegraphics[width=0.5\textwidth]{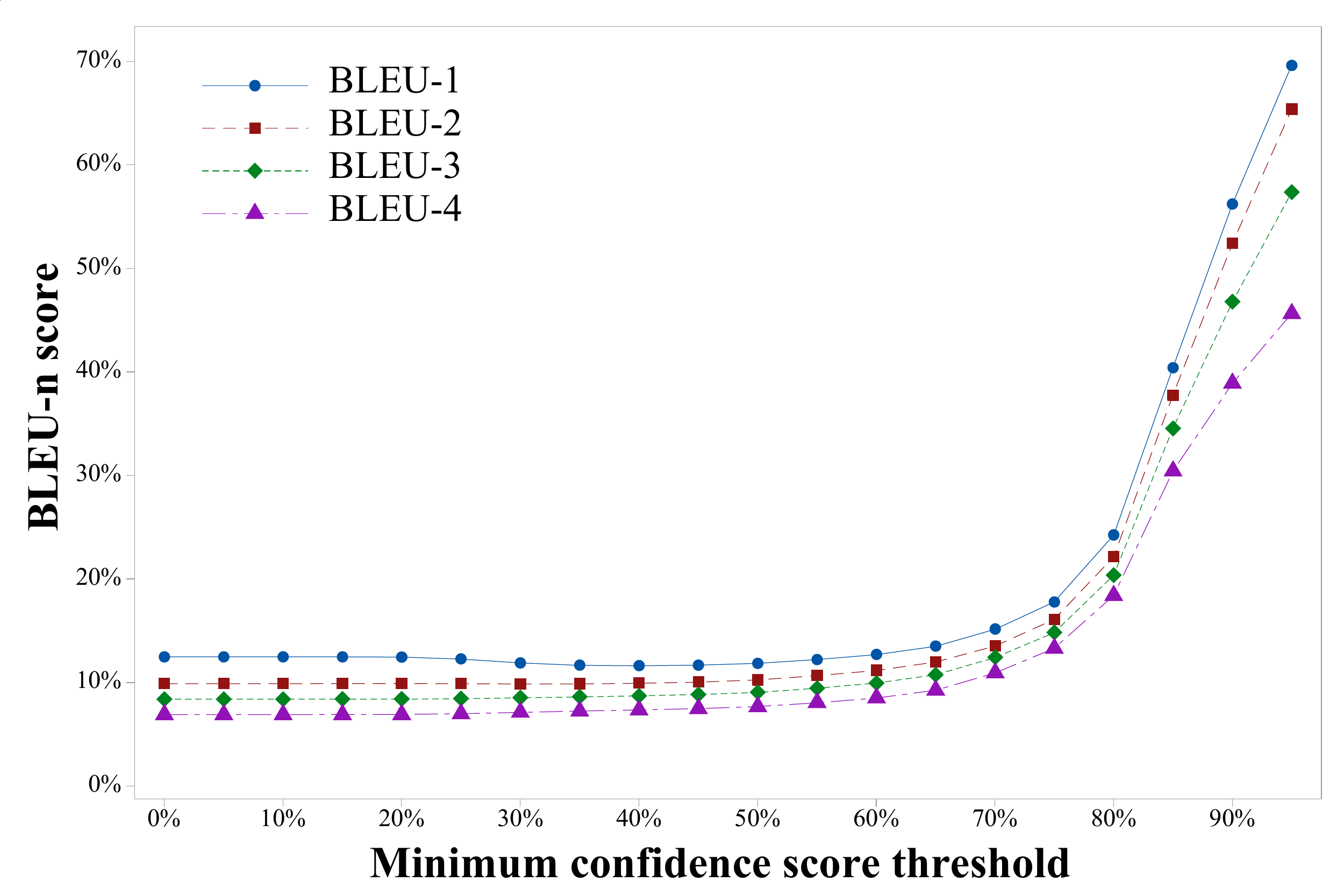}
	\caption{BLEU-n scores for samples above a confidence threshold.\label{fig:bleu}} %TODO
\end{figure}

Using these values, we now create a new experiment to obtain the locally optimal values for \texttt{max\_seq\_len} and \texttt{batch\_sz}, setting them to \{6, 10, 15, 20\} and \{50, 100, 200\}, respectively, while \texttt{embedding\_sz} is set to 1,024, \texttt{iters} to 20,000 and \texttt{lr} to 0.001. The results are shown in Figure~\ref{fig:lstm_exp2}, which indicate that the optimal maximum sequence length is 13 and the optimal batch size is 100.

The BLEU-1 score obtained by our corpus of generated captions is 12.5\%, while the BLEU-4 score is 6.9\%. The latter value is significantly lower than that obtained using the Microsoft COCO dataset~\cite{lin2014microsoft}, 27.7\%. This could be explained by the low quality of many of the captions used. Unconventional style and colloquial language in captions effect the BLEU score because this score is highly dependent on the sequence of selected words. We use a confidence score, as described in Section \ref{sec:evaluation}, to measure the strength of each BLEU value. This allows us to isolate high quality BLEU scores. Figure~\ref{fig:bleu} shows how the BLEU-n scores change with increasing confidence score of the captions. For each point in the figure, we compute the BLEU-n score using only the samples that have a confidence score greater than the value on the X-axis. The generated captions exhibit a confidence level higher than 95\% to obtain a corpus-level BLEU-1 score of 70\% and a BLEU-4 score of 45\%. Evaluation using BLEU-1 score is more appropriate for this study given that captions are generally small and frequent larger-grams are scarce. 
%\rob{The figure shows that the generated captions must have a confidence level higher than 95\% to obtain a corpus-level BLEU-1 score of 70\% and a BLEU-4 score of 45\%}
%The figure shows that the generated captions having a corpus-level BLEU-1 score of 70\% have an average confidence as high as 95\%. 

%%%%%%%%%%%%%%%%%%%%%%%%%%%%%%%%%%%%%%%%%%%%%%%%%%%%%%%%%%%%%%%%%%%%%%%%%%%%%%%%%%%%%%%%%%%%%%%%%%%%%%%%%%%%%%%%%%%%%%%%
\subsection{Caption evaluation}\label{sec:results_users}
To qualitatively evaluate the generated captions, we asked five participants to manually assign a score to each generated caption for one hundred images. For each image, we presented the participant with five captions generated by the NIC generator (or less, if there were repetitions). Each set of captions was evaluated using a score on a scale of 1 to 10, with 10 meaning that all five captions have terms related to the image and 1 meaning that no caption has any term related to the image. As a baseline, the participants were asked to rate an image captioning with a score of 2 if only one caption out of five contained exactly one term related to the image.

The one hundred images used for this evaluation are the first hundred images that appear in our website\footnote{Available at: \url{https://auto-captioning.herokuapp.com}}. This subset of images contains 40 images labeled by our CNN predictor as \texttt{food}, 17 images labeled as \texttt{drink}, 16 images as \texttt{outside}, 14 images as \texttt{menu}, and 13 images labeled as \texttt{inside}.

Table~\ref{tbl:user_results} presents a summary of the scores provided by each of the participants. In general, the average of these scores is 5.646, and the median is 6 which indicates that, on average, at least half of the captions generated for an image have some terms related to the image content. Based on an overall calculation of all the scores provided by all participants, 73.2\% of the images have a score of 3 or higher. This indicates that 73.2\% of the images have at least one predicted caption containing terms related to the image. This result demonstrates high quality prediction given that there can be a tremendous amount of possible word combinations for captioning. 

\begin{table}[t!]
\centering
\caption{Results of evaluation of generated image captions.}
\label{tbl:user_results}
\begin{tabular}{|c|r|r|r|r|}
\hline
Evaluator & \multicolumn{1}{c|}{Min} & \multicolumn{1}{c|}{Max} & \multicolumn{1}{c|}{Avg} & \multicolumn{1}{c|}{Median} \\ \hline
Participant 1 & 1 & 10 & 6.29 & 6 \\ \hline
Participant 2 & 1 & 10 & 5.01 & 4 \\ \hline
Participant 3 &  1 & 10 & 4.92 & 3 \\ \hline
Participant 4 & 1 & 10 & 5.61 & 6 \\ \hline
Participant 5 & 1 & 10 & 6.40 & 8 \\ \hline
\end{tabular}
\end{table}
%%%%%%%%%%%%%%%%%%%%%%%%%%%%%%%%%%%%%%%%%%%%%%%%%%%%%%%%%%%%%%%%%%%%%%%%%%%%%%%%%%%%%%%%%%%%%%%%%%%%%%%%%%%%%%%%%%%%%%%%

\begin{figure}[t!]
	\centering
	\includegraphics[width=0.5\textwidth]{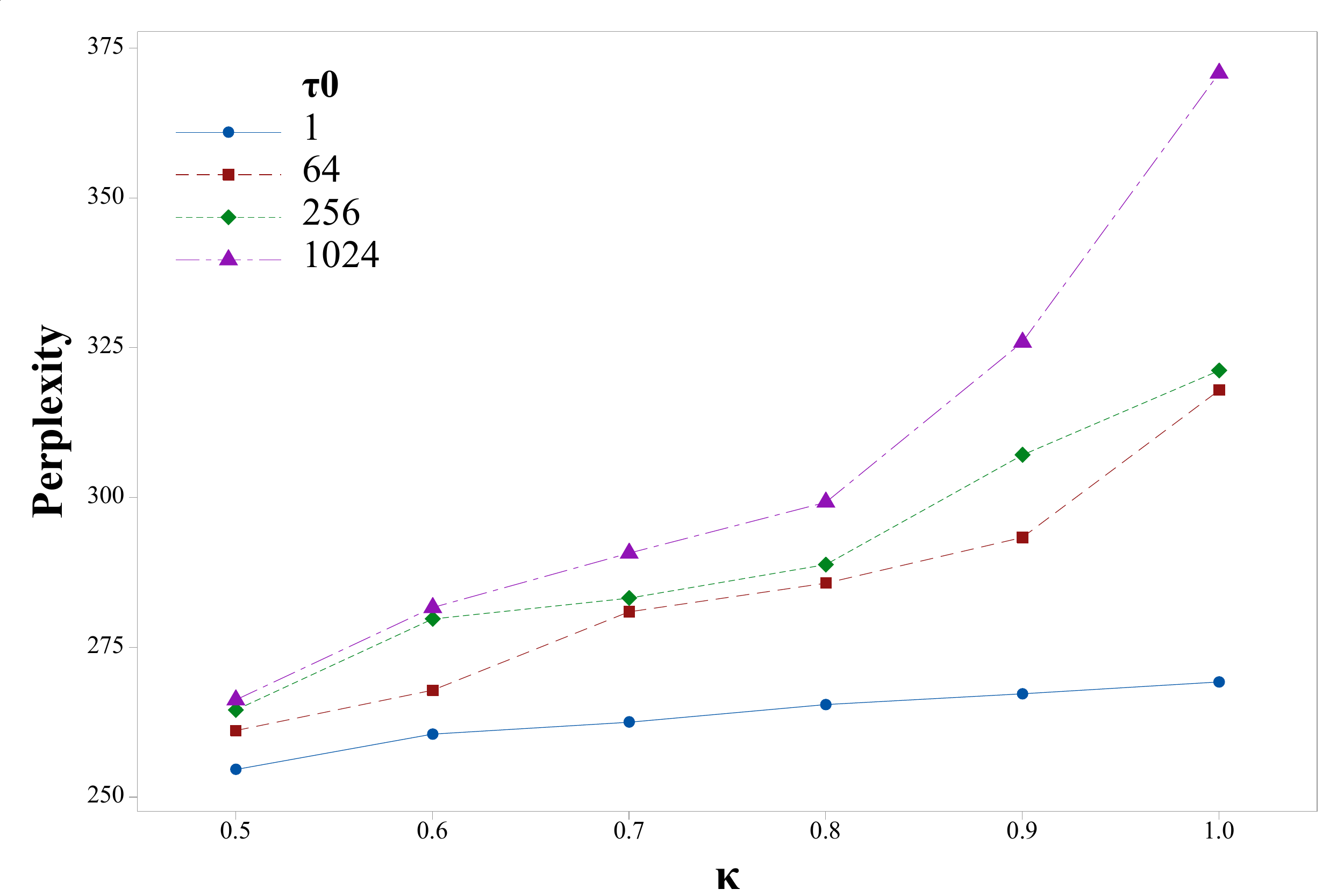}
	\caption{Effect of $\kappa$ and $\tau_0$ settings on LDA model perplexity.\label{fig:lda_exp1}}
\end{figure} %TODO

\subsection{Topic modeling} \label{sec:results_topics}
%\item How does varying different hyperparameters affect the perplexity of the resulting LDA model? (Section~\ref{sec:results_topics}
%\robc{I added the reference to Gensim because if not, the alpha and eta values are meaningless (auto, symmetric, etc.)}
%\rob{The Gensim~\cite{gensim_rehurek_lrec} Latent Dirichlet Allocation implementation used in our framework} 
The Gensim implementation~\cite{gensim_rehurek_lrec} of Latent Dirichlet Allocation used in our framework has the following parameters: number of topics (\texttt{n\_topics}), words per topic, iterations(\texttt{iters}), $\alpha$, $\eta$, $\kappa$ and $\tau_0$. For all of the experiments, we used the ten top words for each topic. As mentioned in Section~\ref{sec:evaluation}, lower perplexity is expected in a better model. In Figure~\ref{fig:lda_exp1}, we show the effect of setting $\kappa$ to \{0.5, 0.6, 0.7, 0.8, 0.9, 1.0\} and $\tau_0$ to \{1, 64, 256, 1024\} on the perplexity of the LDA model, while both $\alpha$ and $\eta$ are set to \textit{symmetric}, \texttt{n\_topics} is 20, and \texttt{iters} is 50. The results show that the locally optimal values for $\kappa$ and $\tau_0$ are 0.5 and 1, respectively.

%In this work, we use the Latent Dirichlet Allocation implementation from Gensim~\cite{gensim_rehurek_lrec}, which has the following hyperparameters: number of topics (\texttt{n\_topics}), words per topic, number of passes, iterations(\texttt{iters}),	$\alpha$, $\eta$, $\kappa$ and $\tau_0$. For all of the experiments the words per topic, or ``top words'' are fixed to 10, while the number of passes is fixed to 8. In Figure~\ref{fig:lda_exp1}, we show the effect of setting $\kappa$ to \{0.5, 0.6, 0.7, 0.8, 0.9, 1.0\} and $\tau_0$ to \{1, 64, 256, 1024\} on the perplexity of the LDA model, while both $\alpha$ and $\eta$ are set to `symmetric', \texttt{n\_topics} is 20, and \texttt{iters} is 50. The results show that the locally optimal values for $\kappa$ and $\tau_0$ are 0.5 and 1, respectively.

\begin{figure}[b!]
 \centering
     \includegraphics[width=0.5\textwidth]{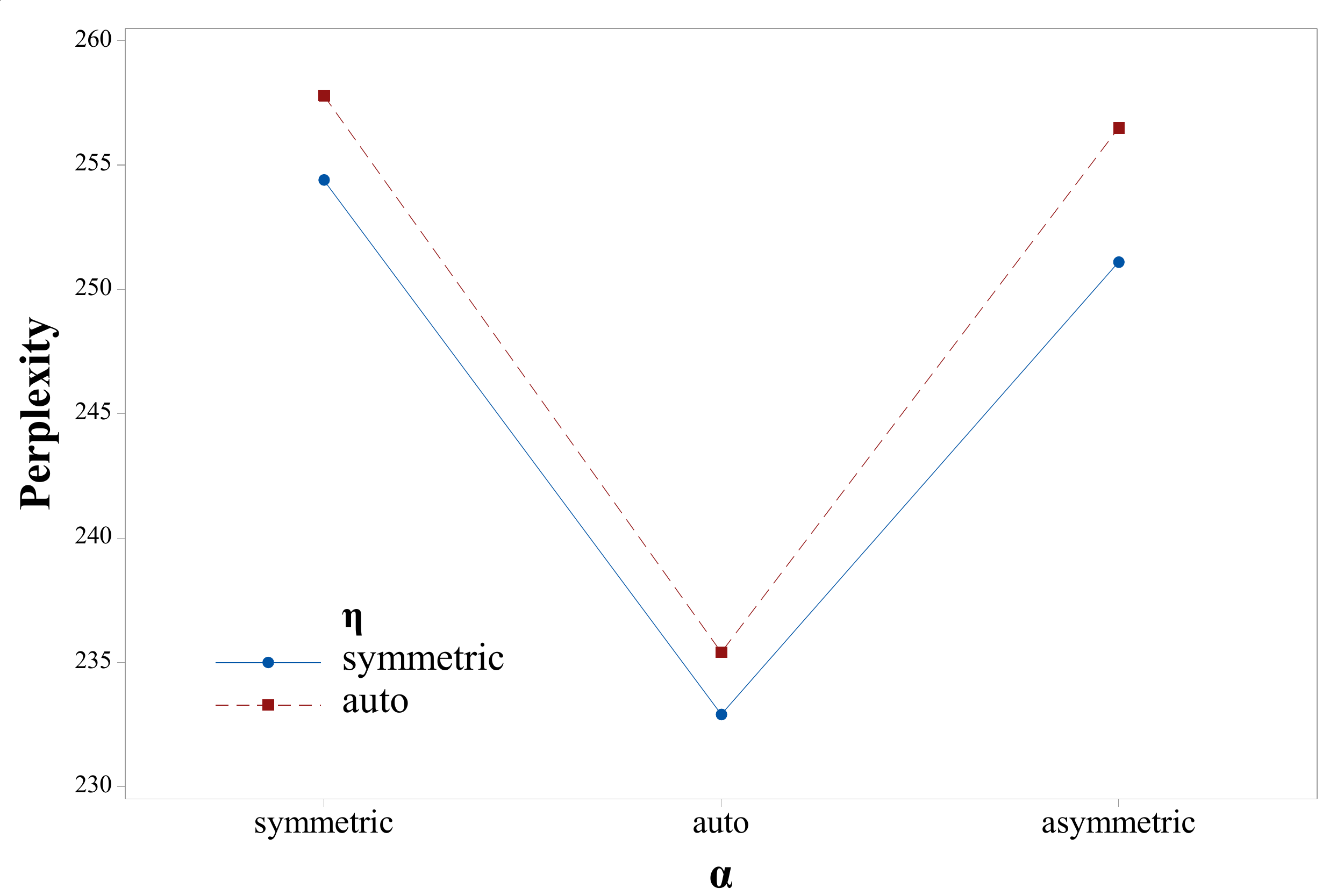}
 \caption{Effect of $\alpha$ and $\eta$ settings on LDA model perplexity.\label{fig:lda_exp2}}
\end{figure}

Using these values for $\kappa$ and $\tau_0$, we test changing $\alpha$ between \{`symmetric', `asymmetric', `auto'\} and $\eta$ between \{`symmetric', `auto'\}, while the other hyperparameters remain the same. Figure~\ref{fig:lda_exp2} presents the effect of changing these variables, where the optimal values for $\alpha$ and $\eta$ are `auto' and `symmetric', respectively.

 %TODO

\begin{figure}[t!]
 \centering
     \includegraphics[width=0.5\textwidth]{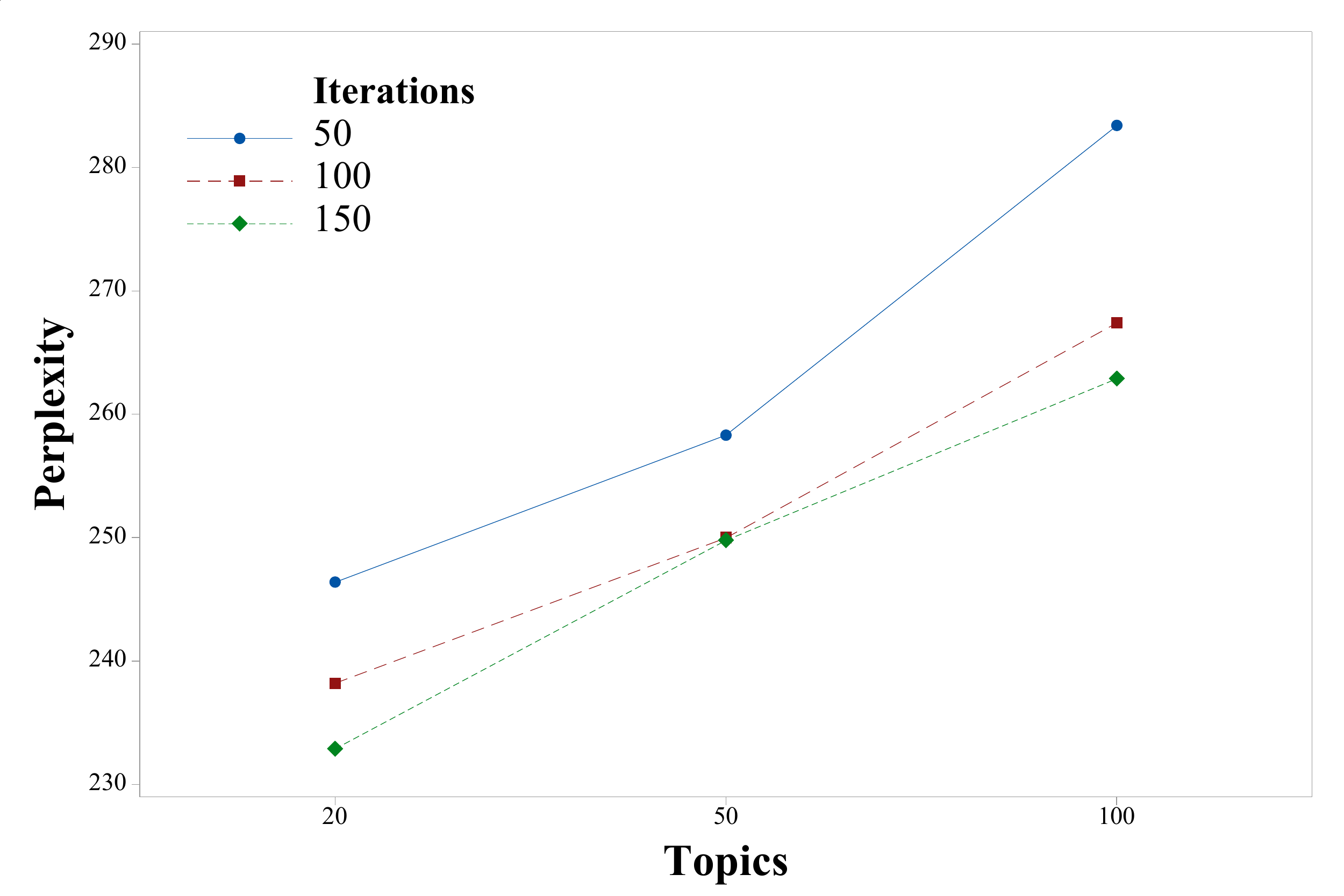}
 \caption{Effect of \texttt{n\_topics} and \texttt{iters} settings on LDA model perplexity.\label{fig:lda_exp3}}
\end{figure}

Finally, we test setting \texttt{n\_topics} to \{20, 50, 100\} and \texttt{iters} to \{50, 100, 150\}, while the other hyperparameters remain the same. Figure~\ref{fig:lda_exp3} presents the effect of changing these variables, where the optimal values for \texttt{n\_topics} and \texttt{iters} are 20 and 150, respectively.

In this section we have shown that certain hyperparameters, such as $\kappa$ and $\tau_0$, can have a significant impact on the perplexity of the model. Our observation is that choosing the wrong values for $\kappa$ and $\tau_0$ may result in a 1.45 times increase in perplexity, while a modification of $\alpha$ and $\eta$ can result in a 1.2 times increase in perplexity. Changes in the number of topics and iterations demonstrate a 1.1 times increase in perplexity. Thus, a careful optimization of these parameters is required to obtain the optimal LDA model in terms of perplexity.

\begin{figure}[b!]
	\fbox{%
		\parbox{0.97\columnwidth}{%
			\textbf{Review:} \textit{We sat on the \textbf{patio} which sits facing the \textbf{Bellagio}. We didn't eat until almost 8PM, the \textbf{fountains} where putting on their show every 15 minutes giving it a little extra to the atmosphere... If you go, and the weather is right, I would recommend sitting on the \textbf{Patio}. The combination of the \textbf{fountains}, the decor and being able to watch people walking up and down the strip made for a fun evening.}
		}%
	}
	\noindent\fbox{%
		\parbox{0.97\columnwidth}{	
			\textbf{Top topical words:} \textit{patio}, \textit{bellagio}, \textit{fountains}, \textit{patio}, \textit{fountains}.
		}%
	}	
	\includegraphics[width=1.0\columnwidth, trim=110 0 120 0, clip]{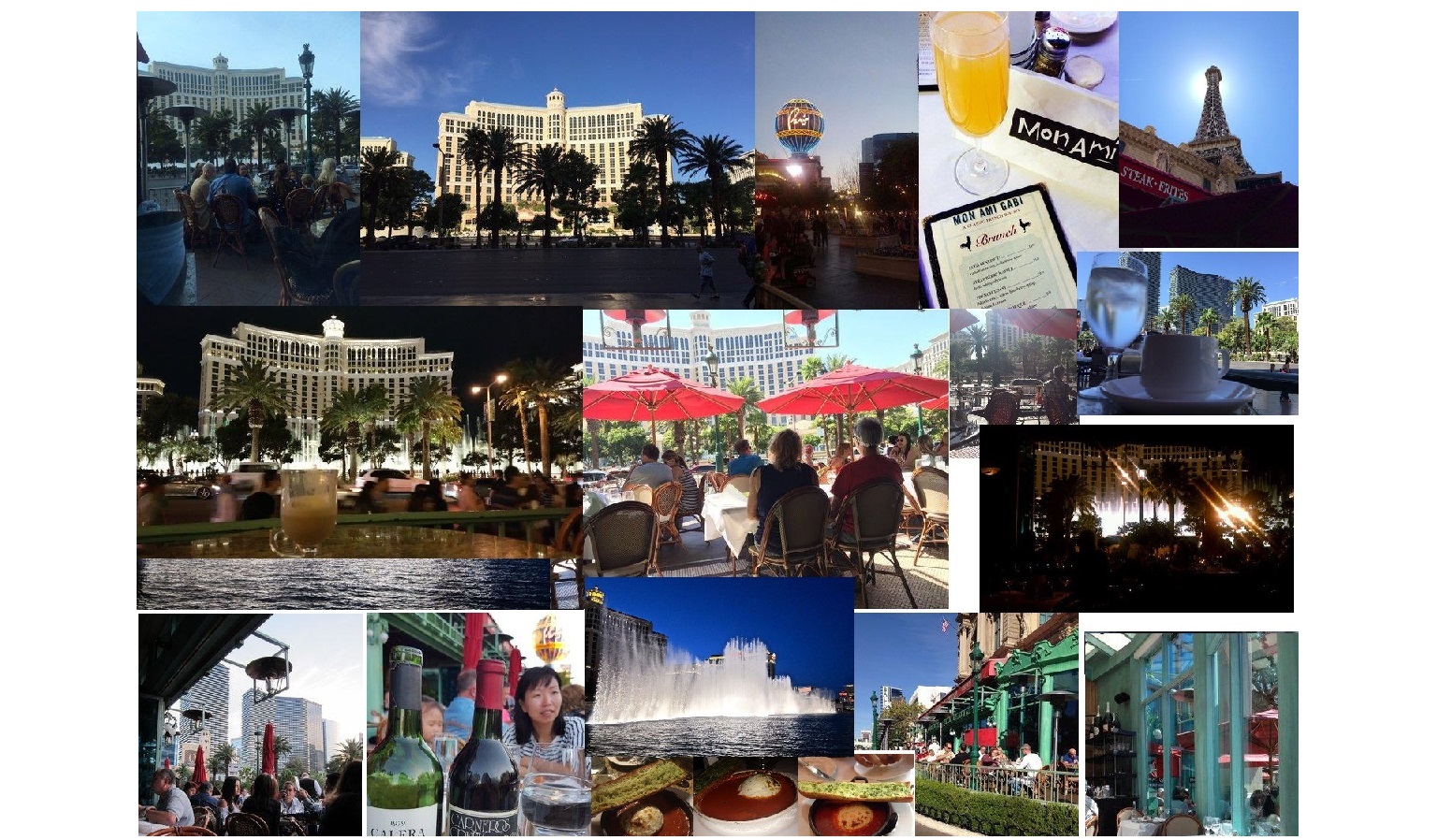}
	\caption{Some of the recommended photos for a review of the Mon Ami Gabi restaurant.}\label{img:recommendedPhotos}
\end{figure}

\subsection{Recommending images for reviews}\label{sec:results_reviews}
%\item Can we relate reviews with images based on the top words obtained through topic modeling?
% To obtain the most probable topic for each review and caption, we start, similar to creating the model, by tokenizing our target review and creating a bag of words using only the indexes of the words. Then, we construct the topic probability distribution from the bag of words. After that, the review is analyzed using the LDA model to obtain the distribution of topics. After obtaining the topics for the reviews and captions, we pair these by matching the terms that they have in common from the top words for the most probable topic. For our results we only used the topic with the highest probability in each document. Each topic was comprised of the 10 most probable words.

% To recommend a photo for a review we go through each image caption and check if there is a word that matches the top words of the review. This process results in a maximum of 10 word comparisons per review and caption instead of comparing each word which is variable.

% For example, the next review had 85\% topic similarity with the captions found to recommend.
In this subsection, we describe a few review-to-image recommendations obtained by our framework. We provided a sample in Figure \ref{img:SuggestedPhotos} of Section \ref{sec:intro} that illustrates that our framework was able to recommend relevant images for a review on a burger-meal. Another review, its top topical words, and recommended images are shown in Figure~\ref{img:recommendedPhotos}. The review was taken from Yelp's entry for the Mon Ami Gabi restaurant in Las Vegas. This example shows that the recommended set contains the images of the fountains of the Bellagio hotel across the street as described in the review. 
 
\begin{table}[t!]
	\caption{Examples of suggested images for reviews of Mon Ami Gabi at Las Vegas.}
	\centering
	\begin{tabular}{|L{3.2cm}|m{4.6cm}|} \hline
		Review summary & Suggested images\\ \hline
		...\textbf{bread} that they serve...recommended the \textbf{french onion soup}...&\includegraphics[width=0.09\textwidth]{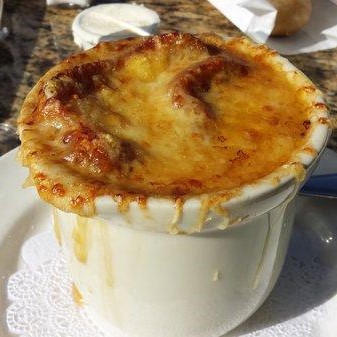}\includegraphics[width=0.09\textwidth]{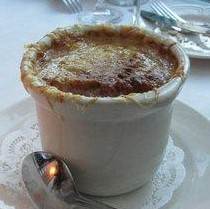}\includegraphics[width=0.09\textwidth]{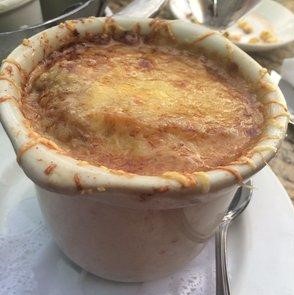}\\ \hline
		...see the \textbf{Bellagio fountain} show...reminiscent of \textbf{Paris}... &\includegraphics[width=0.09\textwidth]{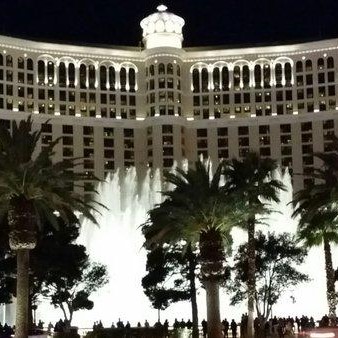}\includegraphics[width=0.09\textwidth]{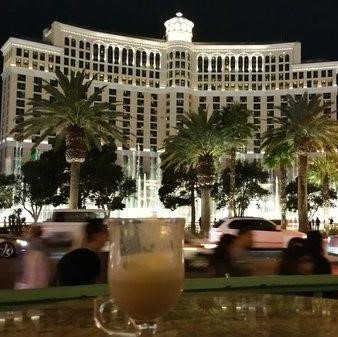}\includegraphics[width=0.09\textwidth]{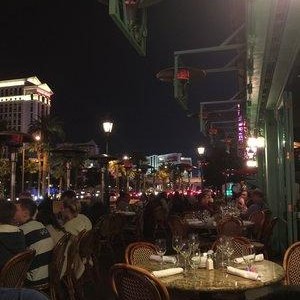}\\ \hline
		...\textbf{Paris} Hotel...who loves \textbf{Paris}...front seat row to the \textbf{Bellagio} light show...&\includegraphics[width=0.09\textwidth]{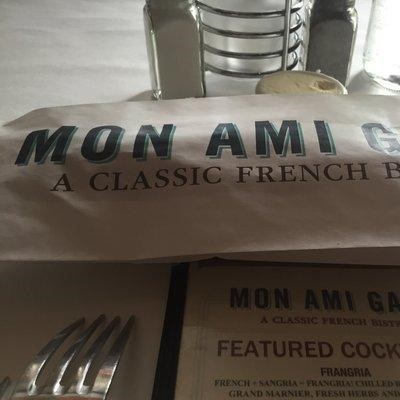}\includegraphics[width=0.09\textwidth]{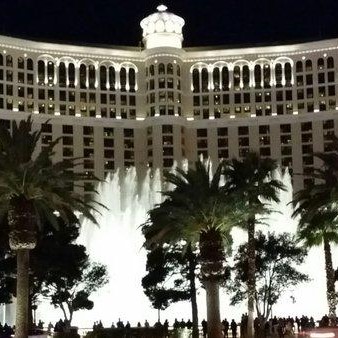}\includegraphics[width=0.09\textwidth]{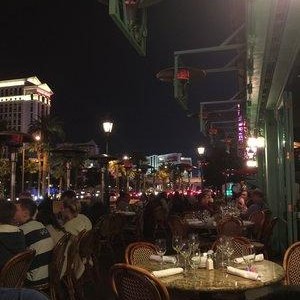}\\ \hline
		...not on the \textbf{menu}... smoked salmon, egg and jack \textbf{cheese}... \textbf{Crepes} were...&\includegraphics[width=0.09\textwidth]{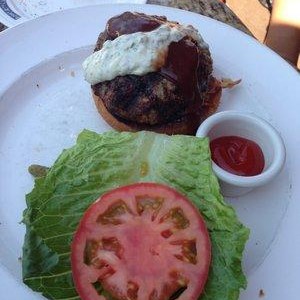}\includegraphics[width=0.09\textwidth]{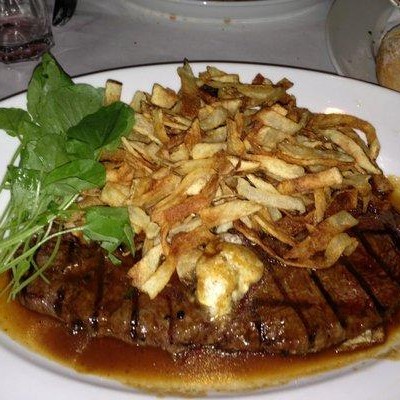}\includegraphics[width=0.09\textwidth]{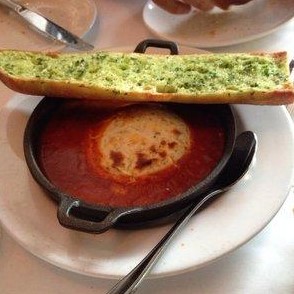}\\ \hline
		...rave about the fabulous \textbf{Bellagio} fountains...traditional \textbf{French menu}...&\includegraphics[width=0.09\textwidth]{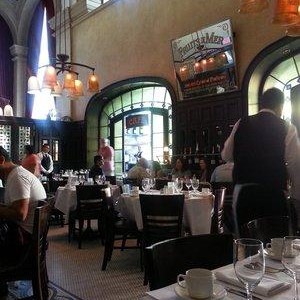}\includegraphics[width=0.09\textwidth]{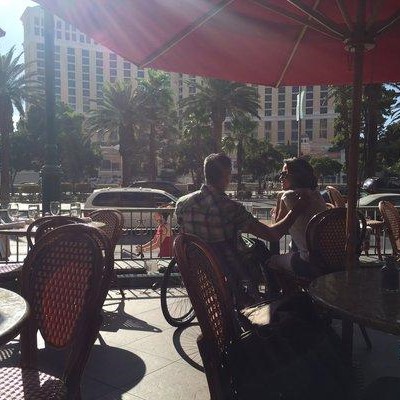}\includegraphics[width=0.09\textwidth]{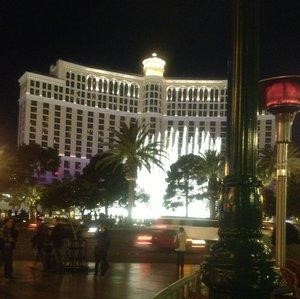}\\ \hline
	\end{tabular}
	\label{tab:review_results}
\end{table} 
 
Summaries of five more reviews with the top three recommended images for each are provided in Table~\ref{tab:review_results}. The table shows that our framework is able to recommend images of main dishes, as well as outside features such as the Bellagio fountains, which are relevant to the reviews. Recommended images for a few thousand reviews of the Mon Ami Gabi restaurant are provided in our website\footnote{\url{https://auto-captioning.herokuapp.com/reviewSuggest.html}}.

%are  To illustrate the relations found by our framework, we present the following example:
%\begin{itemize}
%\item Review: "We sat on the \textbf{patio} which sits facing the \textbf{Bellagio}. We didn't eat until almost 8PM, the \textbf{fountains} where putting on their show every 15 minutes giving it a little extra to the atmosphere...
%If you go, and the weather is right, I would recommend sitting on the \textbf{Patio}. The combination of the \textbf{fountains}, the decor and being able to watch people walking up and down the strip made for a fun evening."
%\item Topic Words found in the review: [`patio', `bellagio', `fountains', `patio', `fountains']
%\end{itemize}
%Figure~\ref{img:recommendedPhotos} shows some of the recommended photos for this review. 

%\rob{Furthermore, in Table~\ref{tab:review_results} we present brief review summaries, with the representative terms in bold, along with the suggested image for each review. The complete results for the Mon Ami Gabi restaurant in Las Vegas can be found in our webpage.}
%\begin{figure}[ht]
% \centering
%     \includegraphics[width=0.5\textwidth]{images/review1Photo}
% \caption{Some of the recommended photos for the example review from Mon Ami Gabi}\label{img:recommendedPhotos}
%\end{figure}

%%%%%%%%%%%%%%%%%%%%%%%%%%%%%%%%%%%%%%%%%%%%%%%%%%%%%%%%%%%%%%%%%%%%%%%%%%%%%%%%%%%%%%%%%%%%%%%%%%%%%%%%%%%%%%%%%%%%%%%%
% Related literature: The background of the work and what are other similar approaches
\section{Related Work} \label{sec:related_work}
Yelp introduced images in the Yelp Challenge recently. To the best of our knowledge, no previous publications have focused on enhancing Yelp reviews by recommending related images. The literature associated with the tasks involved in this paper is described below.

The problem of image classification has been studied for at least half a century, with initial approaches focusing on manual extraction of textural features~\cite{haralick1974}. Due to the difficulty of manual feature extraction, several automatic algorithms have been developed including histogram-based SVM classification \cite{chapelle1999}, as well as pyramid matching with sparse coding~\cite{5206757} and locality-constrained linear coding~\cite{wang2010}. Other methods use a bag-of-features approach, followed by a classifier such as SVM~\cite{csurka2004visual}.

Recently, there has been a lot of interest in deep neural networks. In the area of image classification, a surge has been observed in convolutional neural networks (CNNs). CNNs are neural networks formed by three different types of layers: convolutional layers, pooling layers and fully-connected layers. These layers can be stacked in many different ways, and research is advancing in the direction of deeper networks. Some of the most relevant CNN models include AlexNet~\cite{NIPS2012_4824}, VGG~\cite{simonyan2014very}, GoogLeNet~\cite{szegedy2015going} and ResNet~\cite{DBLP:journals/corr/HeZRS15}. 

The image captioning algorithm used in this paper, based on the work by Vinyals, et al.~\cite{vinyals2015show}, combines a Convolutional Neural Network (CNN) with a special form of a Recurrent Neural Network (RNN) called Long Short-Term Memory (LSTM). Alternatives to using LSTM include primitive rule-Based systems~\cite{gerber1996knowledge, yao2009image} or object detection combined with templates~\cite{farhadi2010every,li2011composing, kulkarni2013babytalk} or Language Models~\cite{mitchell2012midge, aker2010generating, kuznetsova2012collective, kuznetsova2014treetalk, elliott2013image} for caption generation. These are heavily hand-designed alternatives and would be too laborious to implement. Another alternative method for captioning and ranking of these captions is co-embedding images and text in the same vector space~\cite{socher2014grounded}.

Topic modeling has also been widely used in text mining in the past decade. Of particular interest are latent semantic indexing (LSI)~\cite{deerwester1990indexing} and probabilistic LSI (pLSI)~\cite{Hofmann:1999:PLS:312624.312649}, which map documents to a latent semantic space. Latent Dirichlet Allocation (LDA)~\cite{Blei:2003:LDA:944919.944937}, which is used in this work, is a probabilistic approach that generalizes pLSI. Variations of this algorithm include dynamic topic modeling~\cite{blei2006dynamic} and online LDA~\cite{alsumait2008}. Neural networks have also been used for topic modeling~\cite{Larochelle2012, Shamanta2015}.

\section{Conclusions} \label{sec:conclusions}
The framework we designed to enhance Yelp reviews by recommending images requires no supervision. The training samples are gathered from the existing information pieces available with the data. A part of the proposed methodology focuses on enhancing and improving the existing data by providing additional information, i.e., categorizing images and predicting caption-words for them. One of the future directions of this work is to provide further enhancements through the use of multi-label classification where existing caption-words will be used as labels. Another future goal is to develop models to track the performance of a business and the sentiment detected in review and caption texts. 
%\rob{Another future goal is to develop models to track a business’ performance and the sentiment detected in review and caption texts.}
%The framework we designed to enhance Yelp reviews with relevant images could be linked to the actual Yelp website and it might, up to certain degree, enhance the Yelp user's experience. The image classification model and LDA model provide very good results. However we acknowledge that our framework is not flawless. In particular, the very nature of Yelp reviews and image captions means that the content does not necessarily reflect reality, which could result in a reduction of the accuracy of our framework. This, in turn, can translate into suggestions that are irrelevant to the user, which would definitely defeat the purpose of improving the user's experience. 

%%%%%%%%%%%%%%%%%%%%%%%%%%%%%%%%%%%%%%%%%%%%%%%%%%%%%%%%%%%%%%%%%%%%%%%%%%%%%%%%%%%%%%%%%%%%%%%%%%%%%%%%%%%%%%%%%%%%%%%%
\bibliographystyle{abbrv}

\end{document}